\documentclass[lettersize,journal]{IEEEtran}
\usepackage{amsmath,amsfonts}
\usepackage{algorithmic}
\usepackage{array}
\usepackage[caption=false,font=normalsize,labelfont=sf,textfont=sf]{subfig}
\usepackage{textcomp}
\usepackage{stfloats}
\usepackage{url}
\usepackage{verbatim}
\usepackage{graphicx}
\def\BibTeX{{\rm B\kern-.05em{\sc i\kern-.025em b}\kern-.08em
    T\kern-.1667em\lower.7ex\hbox{E}\kern-.125emX}}
\usepackage{balance}

\usepackage{algorithm}
\usepackage{amsmath,amsfonts,bm}

\DeclareMathOperator*{\argmin}{arg\,min}
\usepackage{bbm}
\usepackage{amssymb}
\usepackage{caption}
\usepackage{array,multirow}
\usepackage{adjustbox}
\usepackage{extarrows}
\usepackage{makecell}

\begin{document}
\title{K-means Derived Unsupervised Feature Selection using Improved ADMM}

\author{  Ziheng Sun, Chris Ding, and Jicong Fan
  \thanks{Ziheng Sun, Chris Ding, and Jicong Fan are with the School of Data Science, The Chinese University of Hong Kong (Shenzhen), China. Ziheng Sun and Jicong Fan are also with Shenzhen Research Institute of Big Data, China. Email:
  \texttt{zihengsun@link.cuhk.edu.cn}, \texttt{\{chrisding,fanjicong\}@cuhk.edu.cn}}
}

\markboth{Journal of \LaTeX\ Class Files,~Vol.~18, No.~9, September~2020}%
{How to Use the IEEEtran \LaTeX \ Templates}

\maketitle

\begin{abstract}
Feature selection is important for high-dimensional data analysis and is non-trivial in unsupervised learning problems such as dimensionality reduction and clustering.  The goal of unsupervised feature selection is finding a subset of features such that the data points from different clusters are well separated. This paper presents a novel method called K-means Derived Unsupervised Feature Selection (K-means UFS). Unlike most existing spectral analysis based unsupervised feature selection methods, we select features using the objective of K-means. We develop an alternating direction method of multipliers (ADMM) to solve the NP-hard optimization problem of our K-means UFS model. Extensive experiments on real datasets show that our K-means UFS is more effective than the baselines in selecting features for clustering.
\end{abstract}

\begin{IEEEkeywords}
Feature selection, K-means, ADMM
\end{IEEEkeywords}

\section{Introduction}
\IEEEPARstart{F}{eature} selection aims to select a subset among a large number of features and is particularly useful in dealing with high-dimensional data such as gene data in bioinformatics. The selected features should 
preserve the most important information of the data for downstream tasks such as classification and clustering.
Many unsupervised feature selection methods have been proposed in the past decades. They can be organized into three categories \cite{alelyani2013feature, khalid2014survey}: filter methods \cite{langley1994selection}, wrapper methods \cite{kohavi1997wrappers}, and
hybrid methods. Filter methods evaluate the score of each feature according to certain criteria, such as Laplacian score (LS) \cite{he2005laplacian,zhao2007spectral} and scatter separability criterion \cite{dy2004feature}. Wrapper methods
utilize algorithms to evaluate the quality of selected features.  They repeat selecting a subset of features and evaluating the performance of an algorithm on these features until the performance of the algorithm is desired.  Correlation-based feature selection  \cite{hall1999feature} and Gaussian mixture 
models \cite{constantinopoulos2006bayesian} are representative ones of wrapper methods.
Hybrid methods \cite{das2001filters, liu2005toward} utilize filtering criteria to select feature subsets and evaluate the feature subsets by algorithm performance.

Without labels to evaluate the feature relevance, many criteria have been proposed for unsupervised feature selection in recent years. The most widely used one is to select features that can preserve the data similarity using Laplacian matrix. For instance, Multi-Cluster Feature Selection (MCFS) \cite{cai2010unsupervised} selects features using spectral analysis and 
a regression model with $\ell_1$-norm regularization. Non-negative Discriminative Feature Selection (NDFS) \cite{li2012unsupervised} selects features using non-negative spectral analysis and a regression model with $\ell_{2,1}$-norm regularization. Robust Unsupervised Feature Selection (RUFS) \cite{qian2013robust} selects features using label learning, non-negative spectral analysis and a regression model with $\ell_{2,1}$-norm regularization.
Joint Embedding Learning and Sparse Regression (JELSR) \cite{hou2013joint} selects features using embedding learning and sparse regression jointly. 
Li et al. \cite{li2013extremely} developed a sampling scheme called feature generating machines (FGMs)  to select informative features on extremely high-dimensional problems.
Non-negative Spectral Learning and Sparse Regression-based Dual-graph regularized feature selection (NSSRD)
\cite{shang2017non} extends the framework of joint embedding learning and sparse regression by incorporating a feature graph.
Wang et al. \cite{wang2021autoweighted} proposed to select features using an autoweighted framework based on a similarity graph.
Embedded Unsupervised Feature Selection (EUFS) \cite{wang2015embedded} used sparse regression and spectral analysis. 
Li et al. \cite{li2021sparse} proposed an unsupervised feature selection method based on sparse PCA with 
$\ell_{2,p}$-norm. Sparse and Flexible Projection for Unsupervised Feature
Selection with Optimal Graph (SF$^2$SOG) \cite{wang2022sparse} selects
features using the optimal flexible projections and orthogonal sparse projection with $\ell_{2,0}$-norm constraint.
All these methods select features based on data similarity using spectral analysis, but they don't focus on the separability of data points under the selected feature space. 

In this work, we present a new method called K-means Derived Unsupervised Feature Selection (K-means UFS). Unlike those spectral analysis based methods, we select features to minimize the K-means objective proposed by \cite{ding2004k, zha2001spectral}. The goal of our method is to select the most discriminative features such that the data points are well separated, that is,
have small within-cluster differences and large between-cluster differences. 
We focus on the separability of data points and derive this new unsupervised
feature selection method from K-means. The contributions of this work are as follows.
\begin{itemize}
    \item A novel unsupervised feature selection method is proposed to select the most discriminative features based on the objective of K-means. 
    \item An Alternating Direction Method of Multipliers (ADMM) \cite{boyd2011distributed} algorithm is developed for the NP-hard problem of K-means UFS model.  
    \item We compare K-means UFS with other state-of-the-art unsupervised feature selection methods and conduct experiments on real datasets to demonstrate the effectiveness of our method.
\end{itemize}

 The rest of this paper is organized as follows. In Section \ref{sec_2}, we derive the K-means UFS model from K-means objective. In Section \ref{sec_opt}, we develop an ADMM algorithm to solve the optimization problem of our K-means UFS model. In Section \ref{sec_4}, we compare K-means UFS with other state-of-art unsupervised feature selection methods in detail. Section \ref{sec_5} presents the experiments and Section \ref{sec_conclusion} draws conclusions for this paper.
Table \ref{tab_notation} shows the main notations used in this paper.
\begin{table}
    \centering
    \caption{Notations}\label{tab_notation}
    \begin{tabular}{l|l}
    \hline
        symbol & description\\ \hline
        $\bm{x}$ & column vector  \\
        $\bm{x}_i$ & column vector with index $i$\\
        $X$ & matrix \\
        $\textup{Tr}(\cdot)$ & trace of matrix\\
        $\|\cdot\|_2$ & $\ell_2$ or Euclidean norm of vector\\
        $\|\cdot\|_F$ & Frobenius norm of matrix\\
        $\mathbbm{1}(\cdot)$ & $\mathbbm{1}(x) = 1$ if $x \neq 0$; $\mathbbm{1}(x) = 0$ if $x=0$\\
        \hline 
        $\Vert \cdot \Vert_{2, 1}$ & \makecell{$\ell_{2,1}$ norm of matrix, defined as: \qquad\qquad\qquad\qquad\qquad\qquad\\ $\Vert M \Vert_{2, 1} = \sum_{j=1}^q \sqrt{\sum_{i=1}^p M^2_{ij}} = \sum_{j=1}^q \Vert \bm{m}_j \Vert_2$}\\ \hline
         $\Vert \cdot \Vert_{2, 0}$ & \makecell{$\ell_{2,0}$ norm of matrix, defined as:\qquad\qquad\qquad\qquad\qquad\qquad \\
         $\Vert M \Vert_{2, 0} = \sum_{j=1}^q \mathbbm{1}\Big(\sqrt{\sum_{i=1}^p M^2_{ij}}\Big) = \sum_{j=1}^q  \mathbbm{1}(\Vert \bm{m}_j \Vert_2)$}\\
         \hline
    \end{tabular}
\end{table}

\section{K-means Derived Unsupervised Feature Selection}\label{sec_2}
\begin{figure}[h!]
   \centering
        \includegraphics[width=0.85\columnwidth,trim={0 0 0 5},clip]{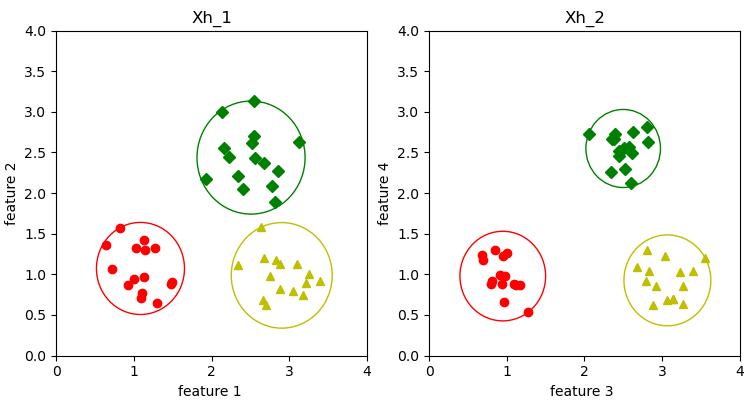}
        \caption{Visualization of $X_{h1}$ and $X_{h2}$ of a toy example}
        \label{fig:diff_X_h}
        \raggedright \small {(Description: The x-axis and y-axis of the left figure represent the first two 
        features (rows) of $X$, while the axis of the right one represent the last two features of $X$.)}
\end{figure}

In unsupervised feature selection, there is no unique criteria to evaluate the quality of selected features. 
We choose to select features by minimizing the objective of K-means clustering proposed by \cite{ding2004k, zha2001spectral}.
Given a data matrix $X = (\bm{x_1}, \bm{x_2}, ..., \bm{x_n}) \in \mathbb{R} ^{p \times n}$,  
where $p$ denotes the number of features and $n$ denotes the number of samples. Each feature (row) of $X$
is standardized to have zero mean and unit variance.  In K-means clustering,  the $k$ centroids are determined by minimizing the sum 
of squared errors, 
\begin{equation}
       \label{eqn:k-means_obj_0}
       \begin{aligned}
           J_k & = \sum_{j=1}^{k} \sum_{i \in C_j} \left\lVert \bm{x_{i}} - \bm{m_j}\right\rVert_2^2,
       \end{aligned}
\end{equation}
where $\bm{m_j} = \sum_{i \in C_j} \bm{x_{i}} / n_j$ is the centroid of cluster $C_j$ consisting of $n_j$ points, $j=1,\ldots,k$.
According to \cite{ding2004k}, $J_k$ can be reformulated as
\begin{equation}
       \label{eqn:k-means_obj}
       \begin{aligned}
           J_k & = \text{Tr}(X^\top X) - \text{Tr}(G^\top X^\top X G),
       \end{aligned}
\end{equation}
where $G = (\bm{g_1}, \dots, \bm{g_k})\in\mathbb{R}^{n\times k}$ is a normalized indicator matrix denoting whether a data point is in a cluster or not, namely, 
\begin{equation}
\label{eqn:indicator_G}
g_{ij}=\left\lbrace
\begin{array}{cc}
1/\sqrt{n_j},& \text{if}~\bm{x}_i\in C_j\\
0,& \text{otherwise}.
\end{array}
\right.
\end{equation}
Let $\Psi $ be the feasible set of all possible indicator matrices with $k$ clusters.  Since $\text{Tr}(X^\top X)$ is a 
constant, \cite{ding2004k} pointed out that K-means clustering is equivalent to
\begin{equation}
    \label{eqn:k-means_problem}
    \begin{aligned}
        \min_{G \in \Psi} ~ - \text{Tr}(G^\top X^\top X G). 
    \end{aligned}
\end{equation}
Our K-means derived unsupervised feature selection (K-means UFS) seeks 
to select the most discriminative features.
The input data $X$ contains $p$ rows of features.
Let  $X_h \in \mathbb{R} ^{h \times n}$ be the selected $h$ rows of $X$.
Our feature selection goals is the following:
{\bf Among all possible choice of $X_h$, we selection the $X_h$ which minimizes the K-means objective}.
Therefore, K-means UFS solves the following problem
\begin{equation}
    \label{eqn:feature_selection_for_k-means_problem_1}
    \begin{aligned}
        \min_{X_h} ~ [\min_{G \in \Psi} ~- \text{Tr}(G^\top X_h^\top X_h G)]
    \end{aligned}
\end{equation}
In order to show the intuition of model \eqref{eqn:feature_selection_for_k-means_problem_1}, we generate a toy data matrix 
$X \in \mathbb{R} ^{4 \times 30}$, shown in Figure \ref{fig:diff_X_h}. The left plot shows $X_{h1}$ consisting of the first two rows of $X$,  while the right 
one shows $X_{h2}$ consisting of the last two rows of $X$. We prefer $X_{h2}$ because the within-cluster differences are much smaller than those in $X_{h1}$.
It is expected that solving problem \eqref{eqn:feature_selection_for_k-means_problem_1} can select the last two features. We want to select the most discriminative features in an unsupervised manner.

\subsection{K-means UFS model}\label{sec_3}
In this section, we show that problem \eqref{eqn:feature_selection_for_k-means_problem_1} can be formulated into a discrete quadratic optimization problem. Let $S \in \mathbb{R} ^{p \times h}$ be a selection matrix defined as:
\begin{equation}
    \begin{aligned}
        X_h = S^\top X,~~\text{s.t.}~ S^\top S=I, ~~ S_{ij} \in \{0, 1\}
    \end{aligned}
    \label{eqn:define_S}
\end{equation}
Let $\Phi$ be the feasible set of all the selection matrix $S$, 
we rewrite problem \eqref{eqn:feature_selection_for_k-means_problem_1} as:
\begin{equation}
    \label{eqn:feature_selection_for_k-means_problem_2}
    \begin{aligned}
        \min_{G \in \Psi,S\in\Phi}- \text{Tr}(S^\top X G G^\top X^\top S)
    \end{aligned}
\end{equation}
Therefore
\eqref{eqn:feature_selection_for_k-means_problem_2} solves both the K-means clustering problem (the optimal $G$) and the feature selection problem (optimal $S$) simultaneously. 

Problem \eqref{eqn:feature_selection_for_k-means_problem_2}
is difficult to optimize. Fortunately, the approximate solution to K-means clustering was obtained in \cite{ding2004k, zha2001spectral}. The approximate solution of K-means indicator
$G^*$ can be constructed as the following. 
Let the singular value decomposition (SVD) of $X$ be $X = P \Sigma Q^\top$,  where $P$ and $Q$ denote the left and right singular vectors and the singular values in $\Sigma$ are sorted decreasingly.  Let $P_k$ be the first 
$k$ columns of $P$, $Q_k$ be the first $k$ columns of $Q$ and $\Sigma_k$ be the first $k$ singular values of $\Sigma$. \cite{ding2004k, zha2001spectral} showed that $Q_k$ is a good approximation of $G^*$. 

Now, with this approximate optimal solution for $G$, we rewrite the objective of problem \eqref{eqn:feature_selection_for_k-means_problem_2} as 
$$\min_{S\in\Phi}~ - \text{Tr}(S^\top P \Sigma Q^\top Q_k Q_k^\top Q \Sigma P^\top S) $$
then, the feature selection optimization problem \eqref{eqn:feature_selection_for_k-means_problem_2} is simplified to
\begin{equation}
      \min_{S\in\Phi}~  -\text{Tr}(S^\top P_k \Sigma_k^2 P_k^\top S) \equiv -\text{Tr}(S^\top A S)
      \label{eq:1}
\end{equation}
 where 
 $$A = P_k \Sigma_k^2 P_k$$
 is in fact the largest $k$ rank of the square of data covariance matrix. Problem \eqref{eq:1} is our model for K-means UFS.
 
 \subsection{Relaxation}
 Problem \eqref{eq:1} is a discrete optimization problem which is typically NP-hard. To solve this problem we use numerical relaxation.
 From Eq.(6), we infer that $S$ has $h$ nonzero rows. This can be seen as the following. The $p$ rows of $X$ can be reordered such that the selected rows are reshuffled to the top $h$ rows of $X$,  
  \begin{equation*}
      \begin{aligned}
      X = \begin{bmatrix}
X_h\\
X_{-h}
\end{bmatrix}, S = \begin{bmatrix}
I_{h \times h}\\
\bm{0}_{(p-h) \times h}
\end{bmatrix}
      \end{aligned}
\end{equation*}
where $X_{-h}$ represents the rest of rows un-selected, $I_{h \times h}$ is an identity matrix and $\bm{0}_{(p-h) \times h}$ is a zero matrix. Therefore, $S$ has three constraints:
\begin{equation}
      \begin{aligned}
         S^\top S = I,~ \Vert S^\top \Vert_{2,0} = h,~~ S_{ij} \in \{0, 1\}
      \end{aligned}
      \label{eq:2}
\end{equation}

 Now, from the K-means UFS model \eqref{eq:1}, if $S^*$ is an optimal solution, $V=S^* R$ (where $R \in \mathbb{R} ^{h \times h}$ is a rotation matrix, that is, $R R^\top = I$.) is also an optimal solution, because $\text{Tr}(S^\top A S) =  \text{Tr}(R^\top S^\top A S R)$. Thus the binary discrete constraint in Eq.\eqref{eq:2} is not necessary.
 Therefore, the relaxed $V$ has only two constraints:
 \begin{equation}
      \begin{aligned}
         V^\top V = I,~ \Vert V^\top \Vert_{2,0} = h
      \end{aligned}
      \label{eq:3}
\end{equation}
Finally, we solve the K-means UFS model by the following relaxed optimization problem.
\begin{equation}
      \label{eqn:L20_problem}
      \begin{aligned}
        \min_V \quad        & -\text{Tr}(V^\top A V)\\
        \textrm{s.t.} \quad & V^\top V = I,~ \Vert V^\top \Vert_{2,0} = h    \\
      \end{aligned}
\end{equation}
Note that once the optimal solution $V^*$ is obtained, the feature selection matrix $S^*$ is determined uniquely by the index of
the $h$ nonzero rows of $V^*$. The value of rotation matrix $R$ has no contribution to feature selection.

\section{Optimization: An Improved ADMM}\label{sec_opt}
In this section, we elaborate how to solve optimization problem \eqref{eqn:L20_problem} using ADMM. 

\subsection{Vanilla ADMM and its limitation}
We consider an equivalent form of problem \eqref{eqn:L20_problem}:
\begin{equation}
    \label{eqn:L20_problem_ADMM}
    \begin{aligned}
     \min_V\quad         & -\text{Tr}(V^\top A V)\\
     \textrm{s.t.} \quad & V = U, \quad U^\top U = I  \\
                         & V = W, \quad \lVert W^\top \Vert_{2,0} = h.  \\
    \end{aligned}
\end{equation}
Then the augmented Lagrangian function of \eqref{eqn:L20_problem_ADMM} is 
\begin{equation}
    \label{eqn:L20_problem_ADMM_aug_Lagrangian}
    \begin{aligned}
    L_{\mu}(V, U, W) = &-\text{Tr}(V^\top A V) + \frac{\mu }{2} \Vert V - U + \Omega/\mu \Vert^2_F \\
                       &+ \frac{\mu }{2} \Vert V - W + \Gamma/\mu\Vert^2_F + \text{const}
    \end{aligned}
\end{equation}
where $\Omega,\Gamma \in \mathbb{R}^{p \times h} $ are Lagrange multiplier matrices, $\mu > 0$ is
a penalty parameter, $U$ is an orthogonal matrix and $W$ is a row-sparse matrix. 
Then we update the variables alternately \cite{boyd2011distributed}:
\begin{align}
V^{t+1} &= \argmin_V L_{\mu}(V, U^{t}, W^{t}) \label{eqn:admm_update_V}\\
U^{t+1} &= \argmin_U L_{\mu}(V^{t+1}, U, W^{t}) \label{eqn:admm_update_U}\\
W^{t+1} &= \argmin_W L_{\mu}(V^{t+1}, U^{t+1}, W) \label{eqn:admm_update_W}\\
\Omega^{t+1}  &= \Omega^{t} + \mu^\top (V^{t+1} - U^{t+1}) \label{eqn:admm_update_Omega}\\
\Gamma^{t+1} &= \Gamma^{t} + \mu^\top (V^{t+1} - W^{t+1}) \label{eqn:admm_update_Gamma}\\
\mu^{t+1} &= \mu^\top \times \rho , \quad \rho = 1.05. \label{eqn:admm_update_mu}\\
\end{align}
\noindent\textbf{Step 1: Update $V$}\\
Let $B = U^\top - \frac{\Omega^\top}{\mu^\top}$, $C = W^\top - \frac{\Gamma^\top}{\mu^\top}$, after algebra, the update $V$ step \eqref{eqn:admm_update_V} is solving the following problem:
\begin{equation}
        \label{eqn:V_step_problem}
        \min_V\quad -\text{Tr}(V^\top A V) + \frac{\mu^\top }{2} \left\lVert V - B \right\rVert^2_F 
        + \frac{\mu^\top }{2} \left\lVert V - C \right\rVert^2_F 
\end{equation}
Take the derivative of this objective function to be zero, we can update $V$ in $t$ iteration by:
\begin{equation}
    \label{eqn:solution_of_V_step_problem}
    V^{t+1} = \frac{1}{2} \mu^\top(\mu^\top I - A)^{-1}(B + C)
\end{equation}
To guarantee the minimal solution exists, $(\mu^\top I - A)$ should be a positive definite matrix. 
Suppose $\lambda_1$ is the max eigenvalue of $A$, we select the initial value of $\mu$ as 
$\mu^0 = \lambda_1 + 0.1$. 
\noindent\textbf{Step 2: Update $U$}\\
The update $U$ step \eqref{eqn:admm_update_U} is solving the following problem:
\begin{equation*}
    \begin{aligned}
    \min_{U}\quad        &\frac{\mu^\top }{2} \Vert V^{t+1} - U + \Omega^\top/\mu^\top \Vert^2_F\\
    \textrm{s.t.} \quad &U^\top U = I  
    \end{aligned}
\end{equation*}
Let $D = V^{t+1} + \Omega^\top/\mu^\top$, it is equivalent to:
\begin{equation}
    \label{eqn:U_step_orthogonal_problem}
    \begin{aligned}
        \max_U\quad        & \text{Tr}(D^\top U) \\
        \textrm{s.t.} \quad & U^\top U = I  
    \end{aligned}
\end{equation}
Let $P_h$ be the first $h$ column of $P$ from 
$D = P \Sigma Q^\top$ (SVD). The solution is:
\begin{equation}
    \label{alg:ADMM_update_U}
    \begin{aligned}
        U^{t+1} &= P_h Q^\top 
    \end{aligned}
\end{equation}

\noindent\textbf{Step 3: Update $W$}\\
Let $F = V^{t+1} + \Gamma^\top/\mu^\top$, the update $W$ step \eqref{eqn:admm_update_W} is solving the following problem:
\begin{equation}
    \label{eqn:ADMM_MV_W_step_problem}
    \begin{aligned}
        \min_{W}  \quad  &\Vert F - W \Vert^2_F \\
        \textrm{s.t.} \quad &\Vert W^\top \Vert_{2,0} = h   \\
    \end{aligned}
\end{equation}
Now we split $W$ and $F$ by rows:
\begin{align*}
    W^\top &= [\bm{w}^{1T}, \bm{w}^{2T}, \dots, \bm{w}^{pT}] \\
    F^\top &= [\bm{f}^{1T}, \bm{f}^{2T}, \dots, \bm{f}^{pT}]
\end{align*}
Select $\bm{l} = \{l_1, l_2, ..., l_h\}$ as the subset of $h$ row indices in $F$ satisfy:
\begin{align*}
    \Vert \bm{f}^{l_1} \Vert_2 \ge \Vert \bm{f}^{l_2} \Vert_2 \ge \dots \ge \Vert \bm{f}^{l_h} \Vert_2
    \ge \Vert \bm{f}^{j} \Vert_2,~~\forall j \notin \bm{l}
\end{align*}
Thus, we update each row of $W$ in $t$ iteration by:
\begin{equation}
    \label{alg:ADMM_update_W}
    \begin{aligned}
        \bm{w}^{(t+1)j} &= \bm{f}^j, \quad \forall j \in \bm{l} \\
        \bm{w}^{(t+1)j} &= \vec{0}, \quad \forall j \notin \bm{l} \\
    \end{aligned}
\end{equation}

\begin{figure}[h!]
       \begin{center}
           \includegraphics[width=0.8\columnwidth, trim={20 8 35 35},clip]{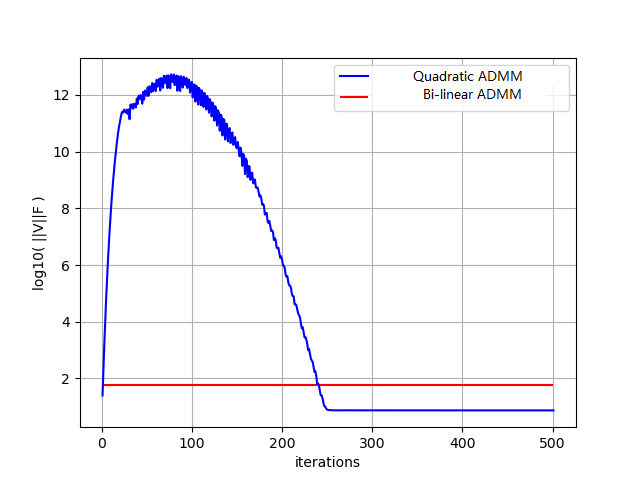}
           \caption{$\log_{10}(\Vert V\Vert_F)$ of Quadratic ADMM Eq. \eqref{eqn:L20_problem_ADMM} and Bi-linear ADMM Eq. \eqref{eqn:L20_problem_ADMM_linear}.}
           \label{fig2:ADMM_log_l1_norm_V_h_8}
        \raggedright \small {(Description: The y-axis represents the
        value of $\log_{10}(\Vert V\Vert_F)$ and the x-axis represents
        the iteration of ADMM. The $\Vert V\Vert_F$ of 
        Quadratic ADMM Eq. \eqref{eqn:L20_problem_ADMM} increases into
        $10^{12.5}$ in 80 iterations.)}
       \end{center}
\end{figure}
In Figure \ref{fig2:ADMM_log_l1_norm_V_h_8}, we apply Quadratic ADMM Eq.\eqref{eqn:L20_problem_ADMM}
on StatLog DNA data\cite{king1995statlog} ($3186$ samples, $180$ features and $3$ classes) and obsverve the $\log_{10}(\Vert V\Vert_F)$ (blue) increases into $12.5$ before
convergence, that means the scale of matrix $V$ is blow-up. The very large values in $V$ will dominate the ADMM process. It's caused by the subproblem \eqref{eqn:admm_update_V} in update $V$ step. The $(\mu I - A)$ matrix is an ill-conditioned matrix
for some data and $\mu$, thus we should {\bf NOT} compute the the inversion of it in Eq.  \eqref{eqn:solution_of_V_step_problem}. In next subsection, two useful tricks are presented to avoid the scale blow-up problem. 

\subsection{Bi-Linear ADMM}
The first trick to avoid scale blow-up is to force $\Vert V \Vert_F$ to
be a constant in each iteration of ADMM. Considering $V$ is an orthogonal matrix in Eq. \eqref{eqn:L20_problem}, the norm of $V$ should satisfy the constraint:
\begin{equation*}
      \Vert V \Vert_F^2 = \text{Tr}(V^\top V) =  \text{Tr}(I_{h \times h}) = h
\end{equation*}
The second trick is to change the quadratic objective function ($-\text{Tr}(V^\top A V)$) into a bi-linear one ($-\text{Tr}(V^\top A U)$). This trick can avoid the matrix inversion in Eq.\eqref{eqn:solution_of_V_step_problem}. Using these two tricks, we obtain an equivalent bi-linear form of Quadratic ADMM Eq.\eqref{eqn:L20_problem_ADMM} as following:
\begin{equation}
       \label{eqn:L20_problem_ADMM_linear}
       \begin{aligned}
           \min_V\quad         & -\text{Tr}(V^\top A U)\\
           \textrm{s.t.} \quad & \Vert V \Vert_F^2 = h, \quad V = U, \quad U^\top U = I \\
                               & V = W, \quad \Vert W^\top \Vert_{2,0} = h \\
       \end{aligned}
\end{equation}
The $\Vert V \Vert_F^2 = h$ is necessary for Bi-linear ADMM, though it is redundant in math. We can see it in the update $V$ step.\\

\noindent\textbf{Step 1: Update $V$}\\
Let $B = U^\top - \frac{\Omega^\top}{\mu^\top}$ and $C = W^{t} - \Gamma^{t}/\mu^\top $, the update $V$ step of problem \eqref{eqn:L20_problem_ADMM_linear} is solving the following problem:
\begin{equation}
    \label{eqn:updata_V_linear}
    \begin{aligned}
        \min_V\quad &-\text{Tr}[(U^{tT} A + \mu^\top B^\top + \mu^\top C^\top) V] \\
        \textrm{s.t.} \quad & \Vert V \Vert_F^2 = h
    \end{aligned}
\end{equation}
Let $D = U^{tT} A + \mu^\top B^\top + \mu^\top C^\top$, the solution is:
\begin{equation}
    \label{eqn:ADMM_MV_solution_of_V}
    V^{t+1} = \frac{\sqrt{h} }{\left\lVert D\right\rVert_F} D
\end{equation}
The $\Vert V \Vert_F^2 = h$ constraint is necessary to guarantee the minimal solution for problem \eqref{eqn:updata_V_linear} exits. In each iteration,
updating $V$ by Eq. \eqref{eqn:ADMM_MV_solution_of_V} will
force $\Vert V\Vert_F^2$ to be a constant $h$. In Figure \ref{fig2:ADMM_log_l1_norm_V_h_8}, the $\log_{10}(\Vert V\Vert_F)$ (red) of Bi-linear ADMM is always a constant. \\
\noindent\textbf{Step 2: Update $U$}\\
Let $E = V^{t+1} + \Omega^\top/\mu^\top$, after 
algebra, the update $U$ step of Bi-linear ADMM is solving the following problem:
\begin{equation}
    \begin{aligned}
    \min_{U}\quad        &-\text{Tr}(V^{(t+1)T} A U) + \frac{\mu^\top}{2}\Vert U - E \Vert^2_F\\
    \textrm{s.t.} \quad &U^\top U = I  
    \end{aligned}
\end{equation}
Let $H = A V^{t+1} + \mu E$, $P_h$ be the first $h$ column of $P$ from 
$ H = P \Sigma Q^\top$ (SVD). The solution is:
\begin{equation}
    \label{alg:ADMM_MV_update_U}
    \begin{aligned}
        U^{t+1} &= P_h Q^\top 
    \end{aligned}
\end{equation}

The update $W$ step is exactly the same as Eq. \eqref{alg:ADMM_update_W}, so
we omit it here. With these update rules, our K-means UFS using Bi-linear ADMM is summarized in Algorithm \ref{alg:ADMM}. 

\begin{algorithm}[h!]
\caption{K-means UFS via Bi-linear ADMM}
\label{alg:ADMM}
\begin{algorithmic}[1]
\REQUIRE Data $X$, number of clusters $k$, number of selected features $h$.
\STATE Initialize $\mu = 0.1$, $\rho = 1.05$. $V, U, W$ is initialized using Eq. \eqref{eqn:initialze_V}. $A$ is initialized using Eq. \eqref{eq:1}.
\STATE Set $t = 0$
\REPEAT
\STATE Update $V$ using Eq.\eqref{eqn:ADMM_MV_solution_of_V}.
\STATE Update $U$ using Eq.\eqref{alg:ADMM_MV_update_U}.
\STATE Update $W$ using Eq.\eqref{alg:ADMM_update_W}.
\STATE Update $\Omega$, $\Gamma$ and $\mu$ using Eq.\eqref{eqn:admm_update_Omega}, Eq.\eqref{eqn:admm_update_Gamma}, Eq.\eqref{eqn:admm_update_mu}.
\UNTIL{Convergence}\\
Select the $h$ features determined by the $h$ nonzero rows in $W$.
\end{algorithmic}
\end{algorithm}

\subsection{Discussion on The Initialization}
$V, U, W$ should be initialized into the same value because of the $V = U$ and $V = W$ constraint. We initialize $V$ by removing the discrete $\ell_{2,0}$-norm constraint in problem \eqref{eqn:L20_problem}:
\begin{equation}
        \label{eqn:initialze_V}
        \min_V \quad -\text{Tr}(V^\top A V)~~~\textrm{s.t.} \quad  V^\top V = I
\end{equation}
Let $P_h$ be the first $h$ column of $P$ from $A = P \Sigma P^\top$ (SVD), then we initialize $V = P_h$. Lagrange multiplier matrices $\Omega = \Gamma = \bm{0}$. $\mu$ is empirically set in the range from $10^{-4}$ to $10^{-1}$ depending on the datasets and is updated by $\mu^{t+1} = \mu^\top \times \rho$ in each iteration. If $\mu$ is larger than $\mu_{max} = 10^7$, we stop updating $\mu$. $\rho$ is empirically set to $1.05$ in our algorithm. 

\subsection{Discussion on The Convergence}
The convergence proof of ADMM can be found in \cite{boyd2011distributed}. 
In ADMM process, $W$ is always a row-sparse matrix with $h$ nonzero rows (Eq. \eqref{alg:ADMM_update_W}) which represents the $h$ selected features.
In $t$ iteration, we can recover a unique feature selection matrix $S^\top$ determined by the $h$ nonzero row indices of $W^\top$ because of $W^\top = V^\top$ and $V^\top = S^\top R$. Our termination criteria is: {\bf If $S^\top$ doesn't change in 30 iterations, we stop the ADMM and output the selected features.} In practice, we set the maximum iteration value as $3000$. In experiment, 
our algorithm converges within 300 iterations for all datasets.

\subsection{Discussion on The Time Complexity}
The time complexity of update $V$ step \eqref{eqn:ADMM_MV_solution_of_V}
involves the computation of $D$ and its Frobenius norm, which is both $\mathcal{O}(np)$. The time complexity of update $U$ \eqref{alg:ADMM_MV_update_U} involves computation of $E$ and its SVD,
which is $\mathcal{O}(np)$ and $\mathcal{O}(n p^2)$. The time complexity of 
update $W$ Eq.\eqref{alg:ADMM_update_W} involves computation of $F$ and sort
the norm of rows in $F$, which is $\mathcal{O}(np)$ and $\mathcal{O}(p \log(p))$. Since $n \gg p$, the time complexity of each iteration is $\mathcal{O}(n p^2)$.

\subsection{Discussion on The Reproducibility}
There are no adjustable parameters in the relaxed K-means UFS model  \eqref{eq:3}. The two parameters $\mu$ and $\rho$ come from the ADMM.
If we set $\rho = 1.05$ and using the initialization settings above. Our 
algorithm is a deterministic algorithm with reproducibility for all the datasets we used.

\section{Other Unsupervised Feature Selection Methods}\label{sec_4}
In this section, we compare K-means UFS with other state-of-the-art unsupervised feature selection (UFS) methods. Most other UFS methods select features using spectral analysis \cite{he2005laplacian, zhao2007spectral} and sparse regression. They tends to select features that can preserve the structure of similarity matrix. Our K-means UFS  \eqref{eqn:feature_selection_for_k-means_problem_1} select features in a totally different way. We select the most discriminative features using the K-means objective which have smaller within-cluster difference and larger between-cluster difference. 

For convenience, we denote original data $X \in \mathbb{R}^{p \times n}$, similarity matrix\footnote{Different methods use different definition of $\tilde{S}$ and details can be found in their papers. Laplacian score use locality projection\cite{he2003locality}.
In NDFS and RUFS, they use the k-nearest similarity matrix $\tilde{S}$. Denote $\mathcal{N}_k(i)$ as the set of k-nearest nodes of $\bm{x_i}$, then $$\tilde{S}_{ij} = \begin{cases}\exp{\Vert \bm{x_i} - \bm{x_j} \Vert^2 / \sigma},~& \bm{x_i} \in \mathcal{N}_k(j)~\text{or}~\bm{x_j} \in \mathcal{N}_k(i)   \\ 0,~&\text{otherwise} \end{cases}$$} 
$\tilde{S} \in \mathbb{R}^{n \times n}$,
$\bm{1} = [1, ..., 1]^\top$, degree matrix $D = \text{diag}(\tilde{S} \bm{1})$, Laplacian matrix $L = D - \tilde{S}$, normalized Laplacian matrix $\hat{L} = D^{-\frac{1}{2}} (D - \tilde{S}) D^{-\frac{1}{2}}$, and $k$ is the number of clusters. Let $Y \in \mathbb{R}^{n \times k}$ be the cluster indicator matrix,
$W \in \mathbb{R}^{p \times k}$ be the regression coefficient weights matrix,
$V \in \mathbb{R}^{p \times k}$ be the latent feature matrix.

\noindent\textbf{Nonnegative Discriminative Feature Selection}\\
NDFS \cite{li2012unsupervised} selects features using non-negative spectral analysis and sparse regression with $\ell_{2,1}$-norm regularization.  They propose the NDFS model as following:
\begin{equation}
\label{eqn:NDFS_reg}
\begin{aligned}
\min_{Y, W}~  & Tr(Y^\top \hat{L} Y) + \alpha \Vert W^\top X - Y^\top \Vert_F^2 + \beta \Vert W^\top \Vert_{2,1} \\
\text{s.t. } & Y^\top Y = I,~~Y \ge 0 \\
\end{aligned}
\end{equation}
where $\alpha$ and $\beta$ are parameters. 

\noindent\textbf{Robust Unsupervised Feature Selection}\\
RUFS\cite{qian2013robust} selects features using non-negative matrix factorization of $X$, non-negative spectral analysis and sparse regression with $\ell_{2,1}$-norm. 
They propose the RUFS model as following:
\begin{equation}
\label{eqn:RUFS_reg}
  \begin{aligned}
  \min_{Y, C, W}\quad  & \Vert X - C Y^\top \Vert_{2,1} + \nu Tr(Y^\top \hat{L} Y) \\
                 & + \alpha \Vert W^\top X - Y^\top \Vert_{2,1} + \beta \Vert W^\top \Vert_{2,1},\\
  \text{s.t. }~~ & Y^\top Y = I,~~Y \ge 0,~~ C \ge 0\\
  \end{aligned}
\end{equation}
where $C \in \mathbb{R}^{p \times k}$ are cluster centers and $\nu$, $\alpha$ $\beta$ are parameters. 

\noindent\textbf{Embedded Unsupervised Feature Selection}\\
EUFS\cite{wang2015embedded} also selects features using non-negative matrix factorization of $X$, non-negative spectral analysis and sparse regression with $\ell_{2,1}$-norm. 
They propose the EUFS model as following:
\begin{equation}
\label{eqn:EUFS_reg}
  \begin{aligned}
  \min_{Y, W}\quad & \Vert X - V Y^\top \Vert_{2,1} + \nu Tr(Y^\top \hat{L} Y)
                 + \beta \Vert V^\top \Vert_{2,1}\\
  \text{s.t. }~~ & Y^\top Y = I,~~Y \ge 0\\
  \end{aligned}
\end{equation}

\noindent\textbf{Sparse and Flexible Projection for Unsupervised Feature
Selection with Optimal Graph}\\
SF$^2$SOG \cite{wang2022sparse} applies spectral analysis on the optimal
graph $G \in \mathcal{R}^{n \times n}$, not similarity matrix $S$. Denote
$L_G \in \mathcal{R}^{n \times n}$ as the normalized Laplacian matrix of 
$G$. They propose the SF$^2$SOG model as follows:
\begin{equation}
\label{eqn:SFSOG_reg}
\begin{aligned}
\min_{Y, W, G}~  & Tr(Y^\top \hat{L}_G Y) + \lambda \Vert W^\top X - Y^\top \Vert_F^2
                    + \gamma \Vert G - S \Vert_F^2 \\
\text{s.t. }  &W^\top W = I, \Vert W^\top \Vert_{2,0} = h, 
              0 \le G_{ij} \le 1, \sum_{j=1}^n G_{ij} = 1 \\
\end{aligned}
\end{equation}

\subsection{Discussion and Comparison}
NDFS \eqref{eqn:NDFS_reg}, RUFS \eqref{eqn:RUFS_reg}, EUFS  \eqref{eqn:EUFS_reg} and SF$^2$SOG \eqref{eqn:SFSOG_reg} are all based on 
the spectral analysis and sparse regression framework. 
Our K-means UFS are totally different from these methods. {\bf First}, K-means UFS selects the most discriminative features based on the K-means objective and focuses on the separability of data points. NDFS, RUFS, EUFS
and SF$^2$SOG consider the local structure of data distribution using spectral analysis and sparse regression. {\bf Second}, the $\ell_{2,0}$-norm constraint used in K-means UFS is derived from the selection matrix $S$ in Eq. \eqref{eqn:define_S} directly. The $\ell_{2,1}$-norm regularization used in NDFS, RUFS and EUFS can be regarded as a relaxation of $\ell_{2,0}$-norm constraints\cite{zhang2014feature}.{\bf Third}, K-means UFS reveals the relationship between data separability and the largest $k$ rank of the square of data covariance matrix $A$ in Eq. \eqref{eq:1}. Researchers can design more advanced methods base on our K-means objective criterion to select features that can increase data separability.

\section{Experiments}\label{sec_5}
In this section, we conduct experiments to evaluate the performance of K-means UFS. Following previous unsupervised feature selection work\cite{zhao2007spectral, li2012unsupervised, cai2010unsupervised}, we only evaluate the performance of K-means UFS in terms of clustering. We evaluate
the performance of clustering using accuracy(ACC) and Normalized Mutual Information(NMI).

\subsection{Datasets}
The experiments are conducted on six real datasets. Some datasets are too large so we 
choose a small subset of them to reduce the time costs of Laplacian matrix computation ($\mathcal{O}(n^2)$) in Laplacian Score, NDFS, RUFS and EUFS. The statistics of the datasets are summarized in Table \ref{table:dataset}.

\begin{itemize}
    \item MicroMass\footnote{MicroMass is in UCI Machine Learning Repository: \url{https://archive.ics.uci.edu/ml/datasets/MicroMass}} is a dataset for the identification of microorganisms from mass-spectrometry data. 
    \item Human Activity Recognition using Smartphones (HARS)\footnote{HARS is in UCI Machine Learning Repository: \url{https://archive.ics.uci.edu/ml/datasets/human+activity+recognition+using+smartphones}} is a dataset for the 
    recognition of human activities using waist-mounted smartphone with embedded inertial sensors. The original dataset consists of 10299 samples so we select the first 500 samples in each classes.
    \item Fashion-MNIST (Fashion)\footnote{Fashion-MNIST is a dataset for Kaggle competitions : \url{https://www.kaggle.com/datasets/zalando-research/fashionmnist}} is a dataset of Zalando's article images consisting of 
    70000 samples including T-shirts, Trousers and so on. We select the first 300 samples in each classes.
    \item Gina\footnote{Gina is a benchmark dataset in IJCNN 2007 Workshop on Agnostic Learning vs. Prior Knowledge: \url{http://www.agnostic.inf.ethz.ch/datasets.php}} is a benchmark dataset for handwritten digit recognition in the agnostic learning.
    \item Fabert and Dilbert\footnote{\url{https://automl.chalearn.org/data}} are benchmark datasets for multiclass tasks in AutoML challenge\cite{guyon2019analysis}. Fabert contains 8237 samepls so we select the first 500 samples in each classes. Dilbert contains 10000 samepls so we select the first 700 samples in each classes.
\end{itemize}

\begin{table}[h!]
    \centering
        \caption{Statistics of the Datasets}
    \label{table:dataset}
    \begin{tabular}{c|c|c|c}
    \hline
    Dataset     & $\#$ of samples & $\#$ of features & $\#$ of classes \\ \hline
    MicroMass   & 1300             &  360              &10               \\ \hline  
    HARS        & 3000              &  561             & 6               \\ \hline
    Fashion     & 3000          &  784            & 10              \\ \hline
    Gina        & 3468             &  784             & 10              \\ \hline 
    Fabert      & 3500            &  800             & 7              \\ \hline
    Dilbert     & 3500           &  2000            & 5               \\ \hline
    \end{tabular}
\end{table}

\subsection{Experimental Settings}
In experiments, the numbers of selected features are set as $\{50, 100, 150, 200, 250, 300\}$ for all datasets.
There are no adjustable parameters in K-means UFS model need to be tuned. 
In experiments, we set $\rho = 1.05$ and use the initialization settings above, the
ADMM (Algorithm \ref{alg:ADMM}) is a deterministic algorithm with reproducibility for 
all the datasets. We compare K-means UFS with the following unsupervised feature selection methods:
\begin{itemize}
    \item[1)]\textbf{All Features:} All original features are adropted.
    \item[2)]\textbf{Laplacian Score} \cite{he2005laplacian}
    \item[3)]\textbf{NDFS:} Non-negative Discriminative Feature Selection  \eqref{eqn:NDFS_reg} \cite{li2012unsupervised}
    \item[4)]\textbf{RUFS:} Robust Unsupervised Feature Selection  \eqref{eqn:RUFS_reg}  \cite{qian2013robust}
    \item[5)]\textbf{EUFS:} Embedded Unsupervised Feature Selection  \eqref{eqn:EUFS_reg}  \cite{wang2015embedded}
    \item[6)]\textbf{SF$^2$SOG:} Sparse and Flexible Projection for Unsupervised Feature Selection with Optimal Graph  \eqref{eqn:SFSOG_reg}  \cite{wang2022sparse}
\end{itemize}
There are some parameters to be set for baseline methods. Following \cite{qian2013robust, wang2015embedded}, we set the neighborhood size to be 5 for the similarity matrix of all
datasets. To fairly compare different unsupervised feature selection methods, the 
$\alpha,\beta,\gamma, \lambda$ parameters in RUFS \eqref{eqn:RUFS_reg}, EUFS \eqref{eqn:EUFS_reg}, SF$^2$SOG \eqref{eqn:SFSOG_reg} are tuned by "grid-search" strategy
from $\{10^{-6}, 10^{-4}, ..., 10^{4}, 10^{6}\}$. We report the best clustering results from 
the optimal parameters. Following \cite{li2012unsupervised, qian2013robust, wang2015embedded}, we use K-means to cluster samples based on the selected features. 
Since K-means depends on initialization, we repeat the clustering 20 times with random
initialization and report the average results with standard deviation.

\begin{table*}[h!]
\scriptsize
\centering
\caption{Clustering Results(ACC $\%$ $\pm$ std) of Different Feature Selection Algorithms}
\label{tab:acc}
\resizebox{\textwidth}{!}{
\setlength{\tabcolsep}{1.2mm}
\renewcommand{\arraystretch}{0.8}
\begin{tabular}{|c|c|c|c|c|c|c|c|}
\hline
Datasets & All Features & Laplacian Score & NDFS & RUFS & EUFS &SF$^2$SOG & K-means UFS \\ \hline
 MicroMass  & 48.1 $\pm$ 4.47 & 49.2 $\pm$ 4.58 & 51.1 $\pm$ 1.75 & 50.6 $\pm$ 3.61 & 50.4 $\pm$ 2.89 & 51.6 $\pm$ 2.73 & \bf 52.3 $\pm$ 5.65 \\ \hline 
 HARS  & 44.6 $\pm$ 0.70 & 46.1 $\pm$ 0.29 & 46.2 $\pm$ 1.15 & 47.2 $\pm$ 0.21 & 46.7 $\pm$ 0.16 & 47.1 $\pm$ 1.32 & \bf 47.4 $\pm$ 0.17 \\ \hline 
 Fashion  & 50.0 $\pm$ 1.46 & 52.1 $\pm$ 1.20 & 53.2 $\pm$ 3.34 & 53.3 $\pm$ 1.88 & 53.8 $\pm$ 2.42 &53.0 $\pm$ 1.24 & \bf 54.1 $\pm$ 1.00 \\ \hline 
 Gina  & 48.0 $\pm$ 1.58 & 46.0 $\pm$ 2.38 & 48.1 $\pm$ 0.14 & 49.5 $\pm$ 2.42 & 46.9 $\pm$ 0.67 & 49.1 $\pm$ 1.25 & \bf 50.1 $\pm$ 0.42 \\ \hline 
 Fabert  & 17.3 $\pm$ 0.51 & 17.1 $\pm$ 0.90 & 17.8 $\pm$ 0.61 & 17.3 $\pm$ 0.72 & 17.3 $\pm$ 0.75 &17.5 $\pm$ 1.32 & \bf 18.0 $\pm$ 0.49 \\ \hline 
 Dilbert  & 35.1 $\pm$ 0.28 & 37.2 $\pm$ 0.39 & 37.9 $\pm$ 0.28 & 37.6 $\pm$ 0.19 & 37.3 $\pm$ 0.09 &37.1 $\pm$ 2.10  & \bf 38.0 $\pm$ 0.57 \\ \hline 
\end{tabular}}
\end{table*}

\begin{table*}[h!]
\scriptsize
\centering
\caption{Clustering Results(NMI $\%$ $\pm$ std) of Different Feature Selection Algorithms}
\label{tab:nmi}
\resizebox{\textwidth}{!}{
\setlength{\tabcolsep}{1.2mm}
\renewcommand{\arraystretch}{0.8}
\begin{tabular}{|c|c|c|c|c|c|c|c|}
\hline
Datasets & All Features & Laplacian Score & NDFS & RUFS & EUFS &SF$^2$SOG & K-means UFS \\ \hline
 MicroMass  & 58.7 $\pm$ 3.53 & 59.3 $\pm$ 3.84 & 61.6 $\pm$ 1.05 & 61.2 $\pm$ 2.75 & 61.3 $\pm$ 1.92 &61.7 $\pm$ 1.11 &\bf 62.3 $\pm$ 5.59 \\ \hline 
 HARS  & 46.6 $\pm$ 0.40 & 47.3 $\pm$ 0.55 & 47.7 $\pm$ 0.82 & 48.1 $\pm$ 0.12 & 48.2 $\pm$ 0.06 &47.1 $\pm$ 1.22 &\bf 48.4 $\pm$ 0.12 \\ \hline 
  Fashion  & 49.7 $\pm$ 0.62 & 50.3 $\pm$ 0.56 & 51.2 $\pm$ 1.90 & 52.1 $\pm$ 0.66 & 51.9 $\pm$ 0.72 &52.2 $\pm$ 1.26 &\bf 52.5 $\pm$ 0.34 \\ \hline 
    Gina  & 43.4 $\pm$ 1.14 & 40.0 $\pm$ 1.22 & 42.2 $\pm$ 0.17 & 44.3 $\pm$ 1.25 & 41.9 $\pm$ 0.42 &43.1 $\pm$ 1.14 &\bf 44.8 $\pm$ 0.34 \\ \hline 
     Fabert  & 2.9 $\pm$ 1.13 & 2.7 $\pm$ 1.12 & 3.4 $\pm$ 0.55 & 3.0 $\pm$ 0.60 & 3.3 $\pm$ 0.98 &3.5 $\pm$ 1.28 &\bf 3.6 $\pm$ 0.77 \\ \hline 
 Dilbert  & 13.6 $\pm$ 0.30 & 14.6 $\pm$ 0.12 & 16.9 $\pm$ 0.24 & 16.4 $\pm$ 0.18 & 15.7 $\pm$ 0.02 &16.1 $\pm$ 1.22 &\bf 17.5 $\pm$ 1.08 \\ \hline 
     
\end{tabular}}
\end{table*}


\subsection{Experimental Results}
We list the experimental results of different methods in Table \ref{tab:acc} and Table \ref{tab:nmi}.
From these two tables, we have the following observations. {\bf First}, feature selection is necessary and effective. 
It can reduce the number of features significantly and improve the clustering performance. 
{\bf Second}, K-means objective criterion can select the most discriminative features such that data points are well
separated. This UFS critierion is effective and is different from other spectral analysis based criteria such as NDFS and RUFS.
{\bf Third}, K-means UFS achieves the best performance for all the datasets we used. 
This can be mainly explained by the following reasons. First, K-means UFS mainly focus on the discriminative information information 
which results in more accurate clustering. Second, \cite{ding2004k, zha2001spectral} pointed out that
the approximate solution $G^*$ \eqref{eq:1} to K-means clustering can preserve the structure information of data $X$. 
Third, the $\ell_{2,0}$-norm constraint in k-means UFS can reduce the redundant and noisy features. 

\begin{figure}[h!]
    \centering
      \subfloat[ACC for MicroMass \\ ($\rho = 1.05$)]{%
        \includegraphics[width=0.48\columnwidth, trim={18 8 30 10},clip]{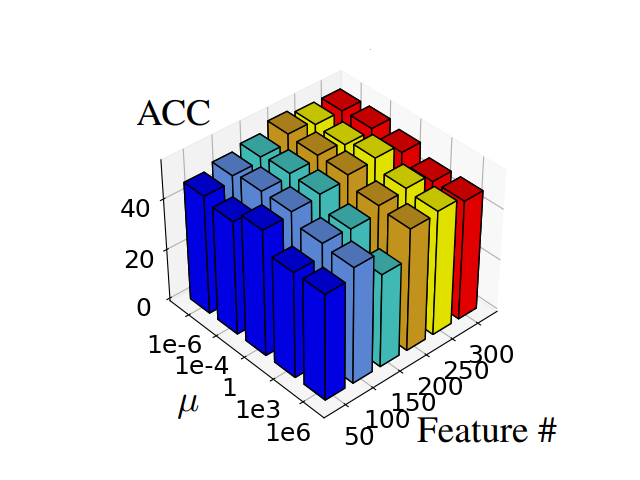}}
      \subfloat[NMI for MicroMass \\($\rho = 1.05$)]{%
      \includegraphics[width=0.48\columnwidth, trim={20 8 30 10},clip]{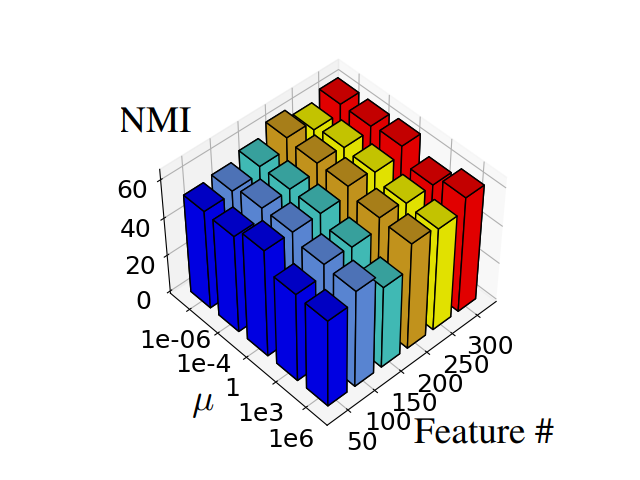}}\\
      \subfloat[ACC for MicroMass \\($\mu^0 = 0.1$)]{%
        \includegraphics[width=0.48\columnwidth, trim={10 8 30 10},clip]{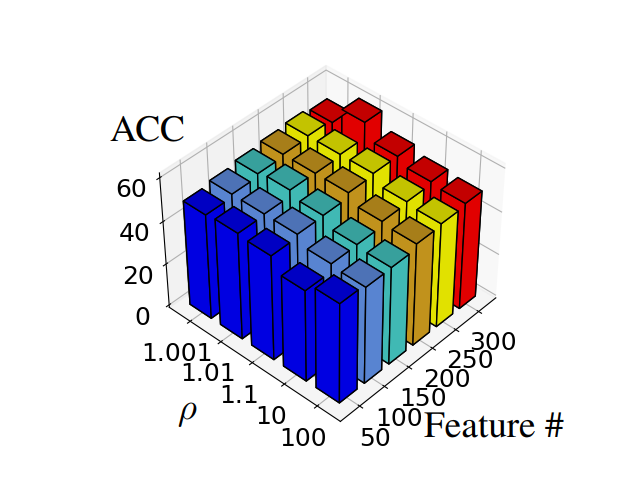}}
      \subfloat[NMI for MicroMass \\($\mu^0 = 0.1$)]{%
        \includegraphics[width=0.48\columnwidth, trim={18 8 30 10},clip]{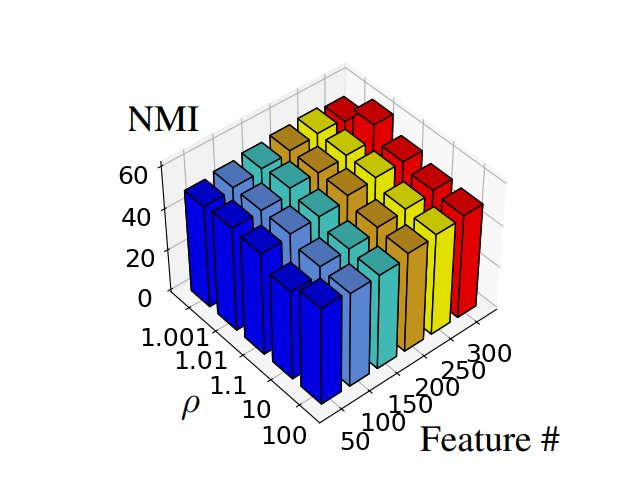}}
    \caption{ACC and NMI of K-means UFS with different initial $\mu^0$ and $\rho$ on MicroMass data}
    \label{fig:param_sens}
    \raggedright \small {(Description: In figure(a)(b), we fix $\rho$ then tune $\mu^0$. In figure(c)(d), we fix $\mu^0$ then tune $\rho$.)}
\end{figure}
\begin{figure}[h!]
    \centering
      \subfloat[MicroMass]{%
        \includegraphics[width=0.48\columnwidth, trim={18 8 30 10},clip]{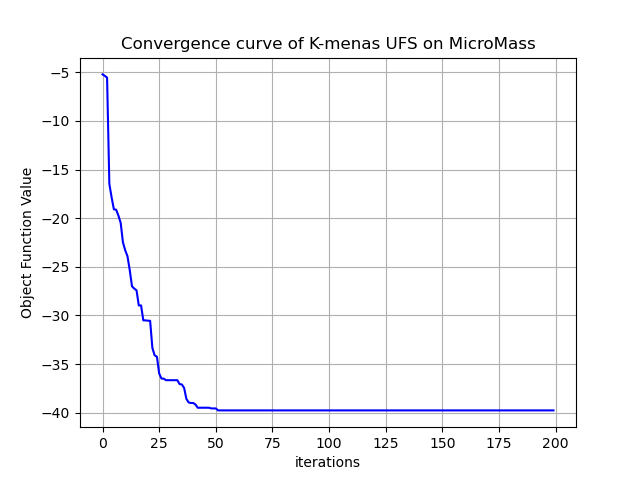}}
      \subfloat[HARS]{%
      \includegraphics[width=0.48\columnwidth, trim={20 8 30 10},clip]{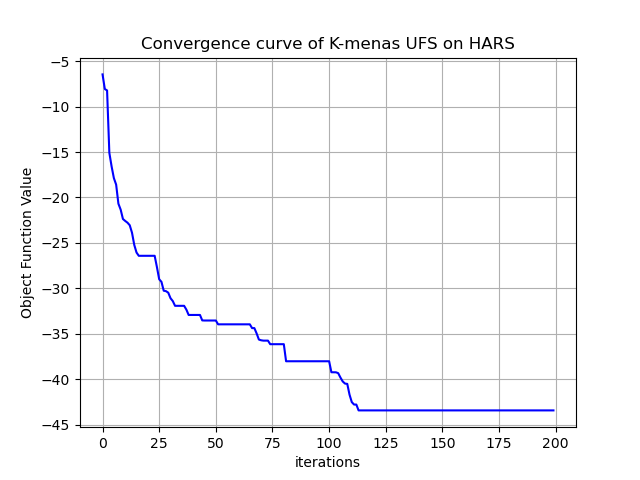}}
    \caption{Convergence curve of K-means UFS on MicroMass and HARS data ($\mu^0 = 0.1, \rho = 1.05, h = 50$)}
        \label{fig:convergence}
        \raggedright \small {(Description: The x-axis represents iterations and the y-axis represents objective value of K-means UFS in ADMM.)}
\end{figure}
We also study the sensitiveness of parameters and the convergence of ADMM. 
There are no adjustable parameters need to be tuned in our K-means UFS model Eq. \eqref{eq:1}. 
We tune the initial value of penalty parameter $\mu^0$ from $\{10^{-6}, 10-^{-3}, ..., 10^6\}$ and its update coefficient $\rho$ from $\{1.0001, 1.01, 1.1, 10, 100\}$ in ADMM (Algorithm \ref{alg:ADMM}).
Due to the space limit, we only report experiments over MicroMass. Fig. \ref{fig:param_sens} show that when 
$\mu^0 \le 1$ and $\rho \le 1.1$, our method is not sensitive to $\mu^0$ and $\rho$. If $\mu^0 \ge 10^3$ or $\rho \ge 10 $, the ADMM will converge within few iterations, which damages the clustering performance. Therefore, we choose $\mu^0 = 0.1$ and $\rho = 1.05$ in experiments. Fig. \ref{fig:convergence} is the convergence curves of K-means UFS over MicroMass and HARS. Our method will converges very quickly and selects
a subset of features with low K-means objective value.

\section{Conclusion}\label{sec_conclusion}
In this paper, we propose a novel unsupervised feature selection method called K-means UFS. 
Our method selects the most discriminative features by minimizing the objective value of K-means.
We derive a solvable K-means UFS model with $\ell_{2,0}$-norm using the approximate indicators of K-means and numerical relaxed trick. We also develop an ADMM algorithm for K-means UFS. The K-means objective criterion
for UFS is totally different from the most widely used spectral analysis criterion. 
Experiments on real data validate the effectiveness of our method.

\bibliographystyle{IEEEtran}
\bibliography{references.bib}

\clearpage

\end{document}